\documentclass{bmvc2k}

\usepackage{pgfplots}

\usepackage{epsfig}
\usepackage{graphicx}
\usepackage{amsmath}
\usepackage{amssymb}

\usepackage{color}
\definecolor{fsublue}{RGB}{62,106,190}
\definecolor{craneblue}{RGB}{4,6,76}

\usepackage{footmisc}
\usepackage{booktabs}
\usepackage{bm}

\usepackage{comment}
\excludecomment{todosection}

\newcommand\cnnout{\bm{f}}

\newcommand\aimg{\bm{z}}
\newcommand\aimgfunc{z}

\newcommand\noiseparam{\bm{\theta}}

\newcommand\transform{\bm{g}}
\newcommand\variance{\mathbb{V}}
\newcommand\expectation{\mathbb{E}}

\newcommand\lcprob{p_{\triangle}}

\newcommand\myparagraph[1]{
\paragraph{#1}
}
\newcommand\sectionname{Sect.}

\newcommand\todo[1]{\textcolor{red}{#1}}

\graphicspath{{figures/imagenoise/}}

\title{Fine-grained Recognition in the Noisy Wild: Sensitivity Analysis of Convolutional Neural Networks Approaches}

\addauthor{Erik Rodner}{erik.rodner@uni-jena.de}{1}
\addauthor{Marcel Simon}{marcel.simon@uni-jena.de}{1}
\addauthor{Robert B. Fisher}{rbf@inf.ed.ac.uk}{2}
\addauthor{Joachim Denzler}{joachim.denzler@uni-jena.de}{1}

\usepackage{url}
\usepackage{color}
\addinstitution{
    Computer Vision Group\\
    Friedrich Schiller University Jena\\
    Germany\\
    \textcolor{blue}{www.inf-cv.uni-jena.de}
}
\addinstitution{
    University of Edinburgh\\
    United Kingdom
}

\def\eg{\emph{e.g}\bmvaOneDot}

\def\etal{\emph{et al}\bmvaOneDot}
\def\ie{\emph{i.e.}\bmvaOneDot}
\runninghead{Rodner \etal}{Sensitivity Analysis of CNNs}

\begin{document}

\maketitle

\begin{abstract}
In this paper, we study the sensitivity of CNN outputs with respect to image transformations and noise in the area of fine-grained recognition.
In particular, we answer the following questions (1) how sensitive are CNNs with respect to image transformations encountered during wild image capture?; (2) how can we predict CNN sensitivity?; and 
(3) can we increase the robustness of CNNs with respect to image degradations?
To answer the first question, we provide an extensive empirical 
sensitivity analysis of commonly used CNN architectures (AlexNet, VGG19, GoogleNet) across various 
types of image degradations.
This allows for predicting CNN performance for new domains comprised by images
of lower quality or captured from a different viewpoint.
We also show how the sensitivity of CNN outputs can be predicted for single images.
Furthermore, we demonstrate that input layer dropout or pre-filtering during test time only reduces CNN sensitivity for high levels of degradation.

Experiments for fine-grained recognition tasks reveal that VGG19 is more robust to severe
image degradations than AlexNet and GoogleNet. However, small intensity noise can lead to dramatic changes
in CNN performance even for VGG19.
\end{abstract}

\section{Introduction}
\label{sec:intro}

    Convolutional neural networks (CNN) are currently the method of choice to model
    a map from visual data to semantic information in applications, such as
     image classification~\cite{krizhevsky2012imagenet,simon2015neural,vgg19}, video categorization~\cite{karpathy2014large}, 
    object detection~\cite{girshick2014rich}, semantic segmentation~\cite{hariharan2015hypercolumns,Brust15:CPN} and many more.
    In contrast to previous approaches with hand-designed feature extraction,
    CNN-based approaches learn relevant features in the form of convolutions directly from the given data.
    
    The driving research question of our paper is how sensitive CNN outputs are
    to image noise and geometric transformations in the area of fine-grained recognition.
    While in nearly every computer vision paper, the evaluation is focused on the expected classification or segmentation accuracy, a more detailed analysis of the sensitivity of the results
    is missing or at least only restricted to the perturbations included in a fixed test set.
    However, being aware of weaknesses of different architectures is crucial to avoid unexpected behavior when the application is deployed. 
    This especially applies to applications where the camera used in the field significantly differs from the camera used to record the training data~\cite{saenko2010adapting}. Fine-grained recognition used for animal monitoring 
    is one example.
    Everyday applications like smartphone apps which use the phone's camera is another example.

    \begin{figure}
        \centering
        \includegraphics[height=0.28\linewidth]{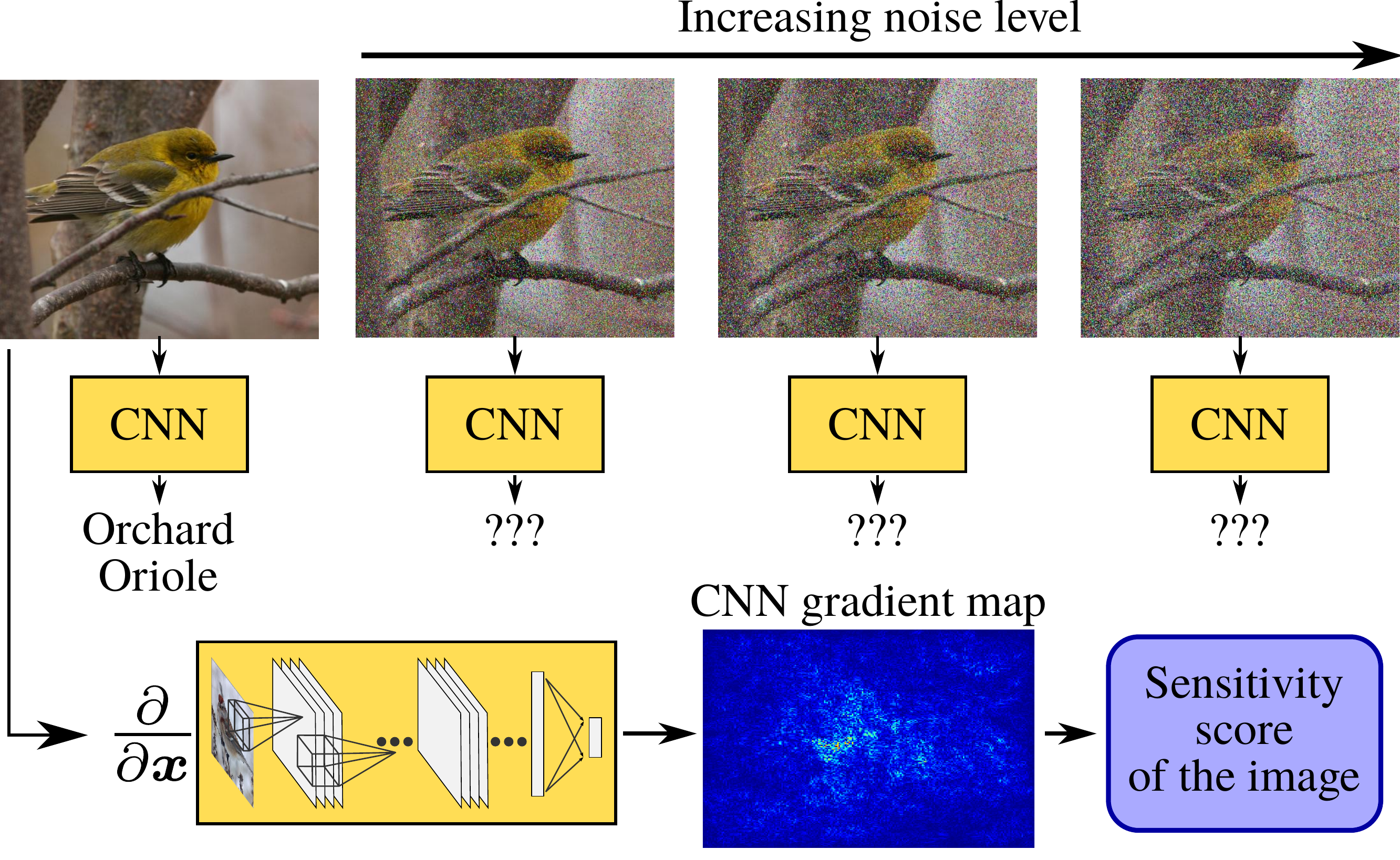}
        \hfill
        \includegraphics[height=0.28\linewidth]{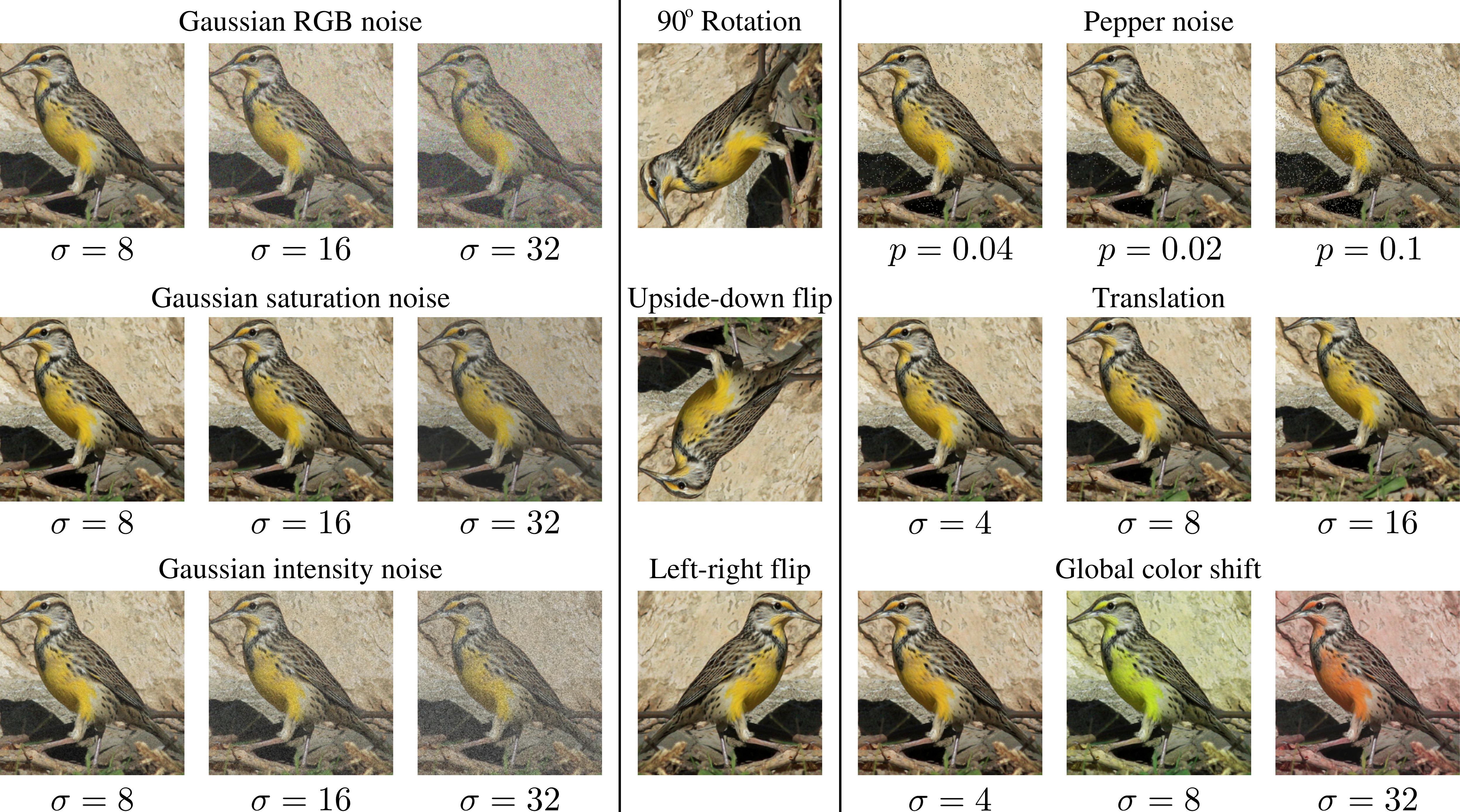}
        \caption{(Left) How sensitive are CNNs with respect to image noise and transformations? We study this question and
        show how to predict CNN sensitivity for a given image. (Right) Examples for the image degradations we use in our paper applied to a single image of the CUB-200-2011 dataset~\cite{CUB200_2011}. This figure is best viewed in color.}
        \label{fig:teaser}
    \end{figure}
    
    \figurename~\ref{fig:teaser} gives an overview of our paper. 
    We provide a sensitivity analysis and prediction approach for common CNN architectures and fine-grained
    recognition.
    First, our analysis allows for selecting networks as well as raising awareness of instabilities
    that might occur when certain perturbations appear more frequently in a new application than in the birth
    place of nearly all pre-trained networks, the ImageNet challenge data~\cite{russakovsky2014imagenet}.
    Second, we show how to compute analytic sensitivity estimates for a given test image without explicitly 
    altering the input image. Our approach is based on a first-order approximation of the CNN outputs and
    can be computed with a single backward pass through the network.
    Third, we show that including a dropout layer directly after the input can significantly boost the
    classification performance for high noise levels.

    Our analysis can also be seen as stochastically studying the ``zone of convergence'' of convolutional neural networks, \ie
    the neighbourhood of an image that leads to the same prediction. Studying these properties of the CNN decision space is important to increase the robustness of visual inference algorithms.

    \myparagraph{Structure of the paper:} The paper is structured as follows: We first review related work in the area of sensitivity analysis.
    An experimental evaluation of CNNs applied to transformed or noisy images is provided in \sectionname~\ref{sec:expsens}.
    \sectionname~\ref{sec:makerobust} shows how to increase the robustness of CNNs with respect to certain degradations
    and \sectionname~\ref{sec:senspred} derives a new algorithm that allows for predicting the sensitivity of a given image.
    The results of the paper are discussed and concluded in \sectionname~\ref{sec:discussion} and \ref{sec:conclusions}.

\subsection{Related work}
\label{sec:relatedwork}

    \myparagraph{Related work on sensitivity analysis of classifiers}
    The sensitivity of standard neural networks (multi-layer perceptrons, MLP) has been studied 
    in the early nineties~\cite{hashem1992sensitivity,choi1992sensitivity} both with respect
    to weight and input perturbations. These works also relate
    the sensitivity to gradient estimates, but are restricted to both standard MLPs and simple vector representations rather than images as in our case.
    A very recent theoretical analysis only limited to linear and quadratic classifiers is given
    in \cite{Fawzi15:FLA}. Furthermore, the study of \cite{Fawzi15:FLA} focuses on adversarial noise rather
    than random noise.

    \myparagraph{Related work on sensitivity analysis of convolutional neural networks}
    The sensitivity of classification systems is related to the concept of adversarial~\cite{szegedy2014intriguing} and 
rubbish examples~\cite{nguyen2014deep}. 
    Adversarial examples are slightly modified images which show a significant change in the model output compared to the original image.
    The work of Szegedy~\cite{szegedy2014intriguing} shows that these alterations can be computed by constrained gradient descent optimization.
    Furthermore, \cite{goodfellow2015explaining} presents the idea of adversarial training, where adversarial examples are added as additional
    training examples. Another strategy to increase the robustness with respect to adversarial examples is given by \cite{gu2014towards}, 
    where contractive networks are proposed that add an additional regularization during training to penalize large gradients
    with respect to the input data. This strategy has been already exploited for autoencoder training previously~\cite{rifai2011higher,rifai2011contractive}. 
    The idea of rubbish examples~\cite{nguyen2014deep} studies random noise images that
    lead to arbitrary classification decisions although their appearance can not be related to the particular object category
    or any natural image at all.
    In contrast to these studies of general robustness, we focus on sensitivity with respect to specific but very common classes
    of image degradations encountered during wild image capture.

    The work of \cite{lenc2014understanding} studies equivariance (with invariance being a special case) and equivalence properties of CNNs
    by explicitly learning transformations of the output to compensate for given degradations. They show that although CNN outputs
    are not invariant to geometric transformations, they are approximately equivarient.
    The paper of \cite{karianakis2015well} analyzes the marginalization properties of CNNs with
    respect to planar translations, scaling, and size of the context around a given bounding box. 
    In contrast to these two works, we mainly focus our analysis on perturbations which occur during the image aquisition process like Gaussian and salt\&pepper noise. 
    Goodfellow~\etal~\cite{goodfellow2009measuring} analyzes the sensitivity in terms of a firing rate, which captures whether a neuron increased or decreased its value above or below a given threshold when pertubations are applied.
    While their analysis is well suited for analyzing the invariance of generic intermediate activations, it is difficult to interpret and understand the results in terms of classification accuracy, which is the task we are interested in. 
    
    \myparagraph{Related work on data augmentation techniques}

    A very natural method to decrease the sensitivity with respect to certain transformations is to
    perform explicit data augmentation by applying perturbations to the training images.
    The VGG19 model~\cite{vgg19}, for example, was trained using color shift data augmentation during training.
    An implicit data augmentation technique is given in \cite{miyato2015distributional}, where an additional regularization
    term is used to minimize the Kullback Leibler divergence between the original and the adversarial posterior distribution.
    In contrast, we study data augmentation and pre-filtering as techniques to explicitly improve CNN performance in the presence of non-adversarial noise.

    \begin{table*}
        \centering
        \resizebox{\linewidth}{!}{
        \begin{tabular}{lcc|cc|cc|cc|cc}
            \toprule
            & \multicolumn{2}{c}{AlexNet} & \multicolumn{2}{c}{VGG} & \multicolumn{2}{c}{GoogLeNet} & \multicolumn{2}{c}{AlexNet} & \multicolumn{2}{c}{AlexNet}\\
            & \multicolumn{2}{c}{CUB200-2011} & \multicolumn{2}{c}{CUB200-2011} & \multicolumn{2}{c}{CUB200-2011} & \multicolumn{2}{c}{Oxford Flowers} & \multicolumn{2}{c}{Oxford Pets}\\
            perturbation & $\lcprob$ & acc.
                         & $\lcprob$ & acc.
                         & $\lcprob$ & acc.
                         & $\lcprob$ & acc.
                         & $\lcprob$ & acc.\\
            \midrule
            no perturbations & 0.00\% & 62.56\% & 0.00\% 	& 81.29\% & 0.00\% & 77.75\%  & 0.00\% & 86.96\%  & 0.00\% & 79.59\%\\
	    \midrule
        $90^{\circ}$ rotation & 60.60\% & 32.22\%  & 51.71\% & 45.93\% & 58.78\% & 38.32\%  & 17.79\% & 78.06\% & 46.44\% & 48.49\%\\
        upside down flip & 65.84\% & 27.06\% & 61.32\% & 36.18\% & 65.08\% & 31.72\%  & 20.59\% & 75.87\% & 56.91\% & 39.17\%\\
        left right flip & 14.05\% & 62.56\% & 10.01\% & 81.15\% & 9.87\% & 77.49\%  & 6.18\% & 86.65\% & 9.89\% & 79.45\%\\
            \bottomrule
        \end{tabular}
           }
        \caption{Label change probability $\lcprob$ and recognition rates for some transformations on CUB-200-2011 and  with different convolutional neural network architectures fine-tuned on the training set of CUB-200-2011.}
        \label{tab:pertub}
    \end{table*}

\section{How sensitive are CNN approaches?}
\label{sec:expsens}

    In this section, we analyze the sensitivity of three state-of-the-art CNN architectures, which are widely used in recent works: AlexNet~\cite{krizhevsky2012imagenet}, VGG19~\cite{vgg19} and GoogLeNet~\cite{googlenet}.
    We show in the experiments the weaknesses of a network that is trained on images which contain almost no noise. 
    This is particularly important in real-world applications, where either low budget cameras are used or the lighting conditions changed after training.

    \myparagraph{Experimental setup}
    We use models pre-trained on the ILSVRC2012 dataset provided by \cite{caffe} and \cite{vgg19}.
    ILSVRC12~\cite{russakovsky2014imagenet} is a large-scale dataset containing roughly 1.5 million training images split into 1000 object categories like car, person, cup, etc.
    Fine-tuning to the application-specific dataset is used, as this is a common step in most tasks. 
    All experiments are performed using the CUB200-2011 birds dataset~\cite{CUB200_2011},
    which is one of the most commonly used datasets in fine-grained recognition.
    It contains 11788 images of 200 north American bird species.
    In addition, we also performed experiments on Oxford Flowers 102~\cite{nilsback08-afc} and the Oxford Pets dataset~\cite{parkhi2012cats}.
    We use the split into training and test provided with the datasets.

    The influence of noise is measured 
    by the label change probability $\lcprob$ and the classification accuracy \textit{acc.}.
    The label change probability is the expected probability that the prediction of the CNN changes if random noise is applied to the image. 
    In other words, if $\lcprob$ is high, the class with the highest predicted probability is likely to change. 
    The classification accuracy measures the performance for the classification task itself, which is related to the label flip probability of the initially correctly classified images.
    
    All images are reduced to the given ground-truth bounding boxes, since we want
    to focus on transformations applied on the objects themselves and not on contextual background.
    Perturbations are applied to resized images 
    fitting the input layer of the CNN. For translations, we translate the bounding box and crop it from the 
    full image to avoid boundary effects.

    \myparagraph{Image noise and geometric transformations}
    We compare several types of noise which occur in real-world applications.
    \figurename~\ref{fig:teaser} (right) shows differently degraded versions of an image illustrating the impact of the noise types and parameters on the visual appearance.

    First, we consider random Gaussian noise affecting different aspects of the image.
    Gaussian RGB noise denotes random additive noise applied to each channel independently with mean 0 and standard deviation $\sigma$, while each pixel in a channel has a value in $[0,255]$.
    Gaussian intensity noise adds the noise to the intensity channel of the HSI image.
    Color shift and saturation noise work on the HSI image as well by adding Gaussian noise to the hue and saturation channel, respectively.
    Global color shift adds the same value to all pixels while local color shift adds a different value to each pixel's hue channel.
    The second type of noise are geometric perturbations of the image.
    We consider random translations, rotations and flips.
    Finally, impulse noise is simulated by randomly setting pixel values to 0, independently for each channel.
    Hence, we call it pepper noise in the figures.

    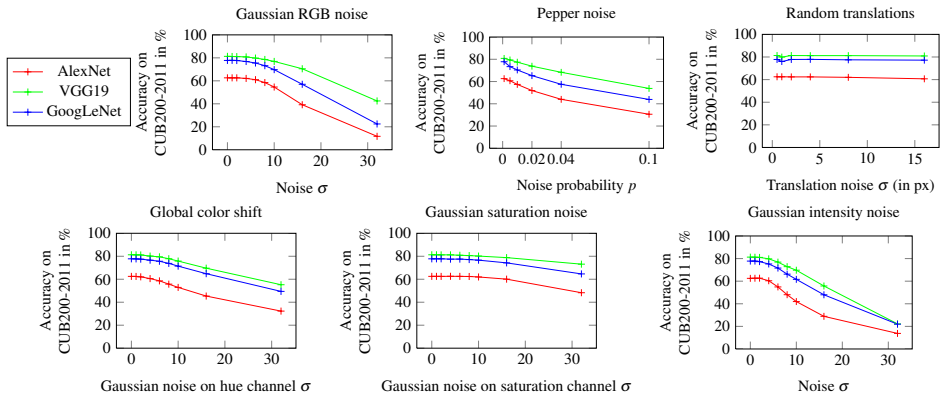
\begin{figure*}[tbp]
  \small
  \centering
      \resizebox{!}{0.2\linewidth}{
        	  \begin{tikzpicture}
	    \begin{axis}[
		    xlabel=Noise $\sigma$,
		    ylabel style={align=center},ylabel=Accuracy on\\ CUB200-2011 in \%,
		    ymin = 0,
		    ymax = 100,
            legend style={at={(-0.8,0.8)},anchor=north},
		    width=0.4\linewidth,
		    height=0.3\linewidth,
		    title=Gaussian RGB noise]
	    \addplot[color=red,mark=+] coordinates {
		    (0, 62.6) (1, 62.6) (2, 62.6)
		    (4, 62.1) (6, 61.0) (8, 58.5)
		    (10, 54.6) (16, 39.2) (32, 11.7)
	    };
	    \addplot[color=green,mark=+] coordinates {
		    (0, 81.3) (1, 81.2) (2, 81.1)
		    (4, 80.8) (6, 79.8) (8, 78.6)
		    (10, 76.9) (16, 70.5) (32, 42.5)
	    };
	    \addplot[color=blue,mark=+] coordinates {
		    (0, 77.75) (1, 77.83) (2, 77.67)
		    (4, 76.82) (6, 75.51) (8, 73.18)
		    (10, 69.75) (16, 56.99) (32, 22.47)
	    };

	    \legend{{AlexNet},{VGG19},{GoogLeNet}}
	    \end{axis}
	  \end{tikzpicture}
      }
      \resizebox{!}{0.2\linewidth}{
        	  \begin{tikzpicture}
	    \begin{axis}[
		    xlabel=Noise probability $p$,
		    ylabel style={align=center},ylabel=Accuracy on\\ CUB200-2011 in \%,
		    ymin = 0,
		    ymax = 100,
            xtick = {0.0, 0.02, 0.04, 0.1},
            xticklabel style={
                    /pgf/number format/fixed,
                    /pgf/number format/precision=5
            },
		    legend pos=south west,
		    width=0.4\linewidth,
		    height=0.3\linewidth,
		    title=Pepper noise]
	    \addplot[color=red,mark=+] coordinates {
		    (0.001, 62.6) (0.005, 60.7) (0.01, 57.4)
		    (0.02, 51.9) (0.04, 44.0) (0.1, 30.6)
	    };
	    \addplot[color=green,mark=+] coordinates {
            (0.001, 80.8) (0.005, 79.4) (0.01, 77.4)
		    (0.02, 73.8) (0.04, 68.3) (0.1, 53.8)
	    };
	    \addplot[color=blue,mark=+] coordinates {
            (0.001, 77.8) (0.005, 73.4) (0.01, 70.4)
		    (0.02, 65.3) (0.04, 57.5) (0.1, 43.9)
	    };

	    \end{axis}
	  \end{tikzpicture}
      }
      \resizebox{!}{0.2\linewidth}{
        	  \begin{tikzpicture}
	    \begin{axis}[
		    xlabel=Translation noise $\sigma$ (in px),
		    ylabel style={align=center},ylabel=Accuracy on\\ CUB200-2011 in \%,
		    ymin = 0,
		    ymax = 100,
		    legend pos=south west,
		    width=0.4\linewidth,
		    height=0.3\linewidth,
		    title=Random translations]
	    \addplot[color=red,mark=+] coordinates {
		    (0.5, 62.5) (1, 62.5) (2, 62.4)
		    (4, 62.4) (8, 62.0) (16, 60.7)
	    };
	    \addplot[color=green,mark=+] coordinates {
            (0.5, 81.3) (1.0, 79.9) (2.0, 81.3)
		    (4, 81.2) (8, 81.2) (16, 80.9)
	    };
	    \addplot[color=blue,mark=+] coordinates {
            (0.5, 77.8) (1.0, 76.0) (2.0, 77.8)
		    (4, 77.9) (8.0, 77.5) (16, 77.2)
	    };

	    \end{axis}
	  \end{tikzpicture}
      }
      \resizebox{!}{0.2\linewidth}{
        	  \begin{tikzpicture}
	    \begin{axis}[
		    xlabel=Gaussian noise on hue channel $\sigma$,
		    ylabel style={align=center},ylabel=Accuracy on\\CUB200-2011 in \%,
		    ymin = 0,
		    ymax = 100,
		    legend pos=south west,
		    width=0.4\linewidth,
		    height=0.3\linewidth,
		    title=Global color shift]
	    \addplot[color=red,mark=+] coordinates {
	    (0,62.50)
(1,62.43)
(2,62.09)
(4,60.58)
(6,58.59)
(8,55.74)
(10,52.89)
(16,45.37)
(32,32.27)

	    };
	    \addplot[color=green,mark=+] coordinates {
(0,81.29) 
(1,81.24) 
(2,81.14) 
(4,80.17) 
(6,79.50) 
(8,77.86) 
(10,75.82)
(16,69.64)
(32,55.23)
	    };
	    \addplot[color=blue,mark=+] coordinates {
	    (0,77.77)
(1,77.72)
(2,77.49)
(4,76.60)
(6,75.65)
(8,73.80)
(10,71.42)
(16,64.84)
(32,49.43)
	    };

	    \end{axis}
	  \end{tikzpicture}
      }
      \resizebox{!}{0.2\linewidth}{
        	  \begin{tikzpicture}
	    \begin{axis}[
		    xlabel=Gaussian noise on saturation channel $\sigma$,
		    ylabel style={align=center},ylabel=Accuracy on\\CUB200-2011 in \%,
		    ymin = 0,
		    ymax = 100,
		    legend pos=south west,
		    width=0.4\linewidth,
		    height=0.3\linewidth,
		    title=Gaussian saturation noise]
	    \addplot[color=red,mark=+] coordinates {
(0,62.50)
(1,62.51)
(2,62.48)
(4,62.57)
(6,62.50)
(8,62.40)
(10,61.98)
(16,60.08)
(32,48.27)
	    };
	    \addplot[color=green,mark=+] coordinates {
	    (0,81.29) 
(1,81.27) 
(2,81.22) 
(4,81.22) 
(6,80.88) 
(8,80.60) 
(10,80.17)
(16,78.82)
(32,73.13)
	    };
	    \addplot[color=blue,mark=+] coordinates {
	    (0,77.77)
(1,77.77)
(2,77.79)
(4,77.49)
(6,77.53)
(8,77.13)
(10,76.61)
(16,74.25)
(32,64.64)
	    };

	    \end{axis}
	  \end{tikzpicture}
      }
      \resizebox{!}{0.2\linewidth}{
        	  \begin{tikzpicture}
	    \begin{axis}[
		    xlabel=Noise $\sigma$,
		    ylabel style={align=center},ylabel=Accuracy on\\CUB200-2011 in \%,
		    ymin = 0,
		    ymax = 100,
		    legend pos=south west,
		    width=0.4\linewidth,
		    height=0.3\linewidth,
		    title=Gaussian intensity noise]
	    \addplot[color=red,mark=+] coordinates {
(0,62.50)
(1,62.60)
(2,62.66)
(4,60.37)
(6,54.91)
(8,48.03)
(10,41.97)
(16,28.89)
(32,13.78)
	    };
	    \addplot[color=green,mark=+] coordinates {
	    (0,81.29) 
(1,81.22) 
(2,81.05) 
(4,79.77) 
(6,76.87) 
(8,72.87) 
(10,69.71)
(16,55.75)
(32,22.21)
	    };
	    \addplot[color=blue,mark=+] coordinates {
	    (0,77.77)
(1,77.77)
(2,77.29)
(4,75.22)
(6,71.59)
(8,66.21)
(10,61.62)
(16,47.96)
(32,21.94)
	    };

	    \end{axis}
	  \end{tikzpicture}
      }
    \caption{Comparison of classification accuracy on CUB200-2011 for Gaussian noise of different types as well as pepper noise and random translations. 
    \label{fig:perturb_plots}}
  \end{figure*}

    \myparagraph{Evaluation}
    The results for CUB200-2011 are given in \figurename~\ref{fig:perturb_plots} for the random noise types as well as in Table~\ref{tab:pertub}
    for the geometric perturbations. 
    The baseline for the classification accuracy is measured on the original test images.

    The rotation as well as the upside down flip cause a fairly high label change probability $\lcprob$ and performance degradation. 
    Interestingly, the classification accuracy is almost unchanged for left right flips and all random translations up to $\sigma = 16px$.
    This is indeed reasonable for our bird recognition task.
    Among random noise, the influence of pepper noise is drastic.
    Setting only 4\% of the image pixels to intensity 0 causes a drop in classification
    performance of $20\%$ for AlexNet and GoogLeNet as well as $10\%$ for VGG19.
    This is surprising since the visual appearance is hardly effected for the human eye (see \figurename~\ref{fig:teaser}, right) for this noise type.
    All Gaussian noise types have a strong influence on accuracy with intensity noise having the most and saturation noise the least influence.
    
\begin{figure*}[tbp]
  \small
  \centering
      \resizebox{!}{0.19\linewidth}{
        	  \begin{tikzpicture}
	    \begin{axis}[
            compat = newest,
		    xlabel=Noise $\sigma$,
		    ylabel style={align=center},ylabel=Accuracy on\\ Oxford Flowers in \%,
		    ymin = 0,
		    ymax = 100,
            legend style={at={(-0.87,1.0)},anchor=north},
		    width=0.4\linewidth,
            ylabel shift=-5,
		    height=0.3\linewidth,
		    title=AlexNet on Oxford Flowers]
	    \addplot[color=green,mark=+] coordinates {
		    (0, 86.96)
(1, 86.92)
(2, 86.88)
(4, 86.50)
(6, 85.23)
(8, 82.83)
(10, 78.50)
(16, 61.39)
(32, 31.04)
	    };
	    \addplot[color=red,mark=+] coordinates {
		    (0, 86.96)
(1, 86.90)
(2, 86.58)
(4, 84.27)
(6, 77.52)
(8, 67.51)
(10, 58.85)
(16, 43.58)
(32, 21.15)
	    };
	    \addplot[color=blue,mark=+] coordinates {
		    (0, 86.96)
(1, 86.90)
(2, 86.85)
(4, 86.69)
(6, 86.38)
(8, 85.64)
(10, 84.69)
(16, 79.05)
(32, 58.64)
	    };
	    \addplot[color=black,mark=o] coordinates {
		    (0, 86.96)
(1, 86.82)
(2, 86.64)
(4, 86.02)
(6, 85.00)
(8, 83.73)
(10, 81.78)
(16, 75.85)
(32, 60.06)
	    };
	    \addplot[color=orange,mark=*] coordinates {
		    (0, 86.96)
(1, 86.88)
(2, 86.48)
(4, 83.56)
(6, 79.81)
(8, 76.68)
(10, 73.56)
(16, 63.52)
(32, 35.75)
	    };

	    \legend{{Gauss. RGB},{Gauss. Intensity},{Gauss. Saturation},{Glo. Color Shift},{Loc. Color Shift}}
	    \end{axis}
	  \end{tikzpicture}
      }
      \resizebox{!}{0.19\linewidth}{
        	  \begin{tikzpicture}
	    \begin{axis}[
            compat = newest,
		    xlabel=Noise $\sigma$,
		    ylabel style={align=center},ylabel=Accuracy on\\ Oxford Pets in \%,
		    ymin = 0,
		    ymax = 100,
		    legend pos=outer north east,
            ylabel shift=-5,
		    width=0.4\linewidth,
		    height=0.3\linewidth,
		    title=AlexNet on Oxford Pets]
	    \addplot[color=green,mark=+] coordinates {
		    (0, 79.59)
(1, 79.51)
(2, 79.37)
(4, 78.85)
(6, 77.91)
(8, 76.71)
(10, 75.23)
(16, 68.43)
(32, 42.56)
	    };
	    \addplot[color=red,mark=+] coordinates {
		    (0, 79.59)
(1, 79.47)
(2, 79.01)
(4, 77.71)
(6, 75.19)
(8, 71.04)
(10, 66.32)
(16, 53.68)
(32, 34.30)
	    };
	    \addplot[color=blue,mark=+] coordinates {
		    (0, 79.59)
(1, 79.58)
(2, 79.53)
(4, 79.36)
(6, 79.17)
(8, 79.00)
(10, 78.58)
(16, 76.82)
(32, 68.40)
	    };
	    \addplot[color=black,mark=o] coordinates {
		    (0, 79.59)
(1, 79.43)
(2, 79.27)
(4, 78.53)
(6, 77.29)
(8, 76.39)
(10, 75.45)
(16, 71.63)
(32, 59.33)
	    };
	    \addplot[color=orange,mark=*] coordinates {
		    (0, 79.59)
(1, 79.52)
(2, 79.48)
(4, 79.20)
(6, 78.69)
(8, 77.87)
(10, 77.44)
(16, 75.03)
(32, 46.07)
	    };

	    \end{axis}
	  \end{tikzpicture}
      }
      \resizebox{!}{0.19\linewidth}{
        	  \begin{tikzpicture}
	    \begin{axis}[
            compat=newest,
		    xlabel=Noise probability $p$,
		    ylabel style={align=center},ylabel=Accuracy on\\ Oxford Flowers/Pets in \%,
		    ymin = 0,
		    ymax = 100,
		    legend pos=south west,
            ylabel shift=-5,
		    width=0.4\linewidth,
		    height=0.3\linewidth,
            xtick = {0.0, 0.02, 0.04, 0.1},
            xticklabel style={
                    /pgf/number format/fixed,
                    /pgf/number format/precision=5
            },
            title=Pepper noise on Oxford Flowers/Pets]
	    \addplot[color=red,mark=+] coordinates {
(0.001, 86.61)
(0.005, 84.22)
(0.01, 79.94)
(0.02, 71.83)
(0.04, 60.47)
(0.1, 41.69) 
	    };
        \addplot[color=blue,mark=o] coordinates {
(0.001, 79.10)
(0.005, 77.56)
(0.01, 75.45)
(0.02, 71.75)
(0.04, 65.47)
(0.1, 53.22) 
	    };

        \legend{{Oxford Flowers},{Oxford Pets}}
	    \end{axis}
	  \end{tikzpicture}
      }
    \caption{Comparison of classification accuracy on Oxford Flowers 102 and the Oxford Pets dataset for a fine-tuned convolutional
    neural network with AlexNet architecture.
    \label{fig:perturb_plots_flowers}}
  \end{figure*}
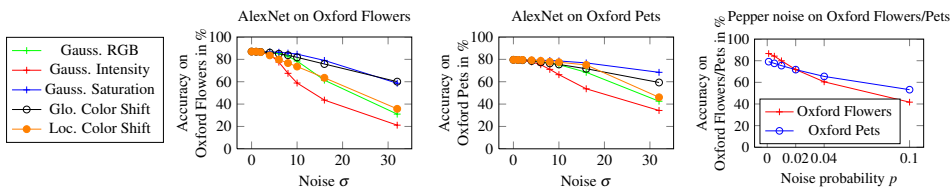

    Fig.~\ref{fig:perturb_plots_flowers} shows results for the Oxford Flowers 102 and the Oxford Pets dataset comparing the different noise types for AlexNet. In summary, similar conclusions compared to CUB200-2011 can be drawn.
    Gaussian intensity noise has by far the strongest impact on accuracy. 
    A noise standard deviation of $\sigma=16$ causes the recognition rate to drop by half to 43.6\% for Oxford Flowers. 
    Noise on the hue channel for every pixel and Gaussian RGB noise have slightly less influence, but are still very noticeable.
    Global color shift and noise on the saturation channel of the images have only little influence. 
    The same is true for flips and rotations, which is reasonable due to flowers being close to rotationally symmetric.

\section{Can we make CNNs more robust?}
\label{sec:makerobust}

    Our experiments showed that even small random noise can lead  to a dramatic performance decrease. 
    Now the question naturally arises whether it is possible to increase the robustness either
    during testing or by adapting the learning.

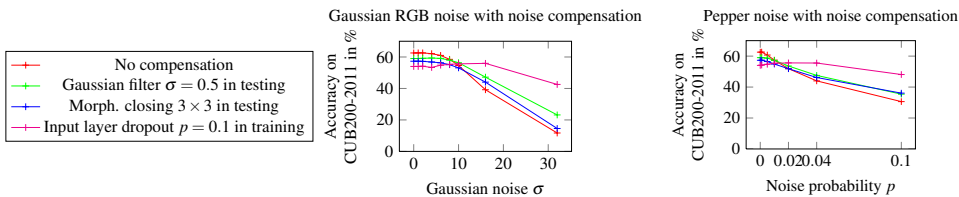
\begin{figure*}
    \centering
    \resizebox{!}{0.20\linewidth}{
        	  \begin{tikzpicture}
	    \begin{axis}[
		    xlabel=Gaussian noise $\sigma$,
		    ylabel style={align=center},ylabel=Accuracy on\\ CUB200-2011 in \%,
		    ymin = 0,
		    ymax = 70,
		    legend style={at={(-0.5,0.5)},anchor=east},
		    width=0.4\linewidth,
		    height=0.3\linewidth,
		    title=Gaussian RGB noise with noise compensation]
	    \addplot[color=red,mark=+] coordinates {
		    (0, 62.56) (1, 62.6) (2, 62.6)
		    (4, 62.1) (6, 61.0) (8, 58.5)
		    (10, 54.6) (16, 39.2) (32, 11.7)
	    };
	    \addplot[color=green,mark=+] coordinates {
		    (0, 59.01) (1, 59.04) (2, 59.15)
		    (4, 59.34) (6, 59.09) (8, 58.01)
		    (10, 56.36) (16, 47.19) (32, 23.18)
	    };
	    \addplot[color=blue,mark=+] coordinates {
		    (0, 57.30) (1, 57.30) (2, 57.22)
		    (4, 56.81) (6, 56.31) (8, 55.11)
		    (10, 53.10) (16, 44.02) (32, 14.56)
	    };
	    \addplot[color=magenta,mark=+] coordinates {
		    (0, 53.99) (1, 53.99) (2, 54.13)
		    (4, 53.35) (6, 54.80) (8, 55.31)
		    (10, 55.63) (16, 55.96) (32, 42.56)
	    };

	    \legend{{No compensation},{Gaussian filter $\sigma=0.5$ in testing},{Morph. closing $3 \times 3$ in testing},{Input layer dropout $p=0.1$ in training}}
	    \end{axis}
	  \end{tikzpicture}
    }    
    \resizebox{!}{0.20\linewidth}{
        	  \begin{tikzpicture}
	    \begin{axis}[
		    xlabel=Noise probability $p$,
		    ylabel style={align=center},ylabel=Accuracy on\\ CUB200-2011 in \%,
		    ymin = 0,
		    ymax = 70,
            xtick = {0.0, 0.02, 0.04, 0.1},
            xticklabel style={
                    /pgf/number format/fixed,
                    /pgf/number format/precision=5
            },
		    width=0.4\linewidth,
		    height=0.3\linewidth,
		    title=Pepper noise with noise compensation]
	    \addplot[color=red,mark=+] coordinates {
            (0.0, 62.56)
		    (0.001, 62.6) (0.005, 60.7) (0.01, 57.4)
		    (0.02, 51.9) (0.04, 44.0) (0.1, 30.6)
	    };
	    \addplot[color=green,mark=+] coordinates {
            (0.0, 59.01)
            (0.001, 59.01) (0.005, 58.73) (0.01, 57.27)
		    (0.02, 53.62) (0.04, 47.53) (0.1, 35.41)
	    };
	    \addplot[color=blue,mark=+] coordinates {
            (0.0, 57.30)
            (0.001, 57.36) (0.005, 56.51) (0.01, 54.87)
		    (0.02, 51.85) (0.04, 46.18) (0.1, 36.07)
	    };
	    \addplot[color=magenta,mark=+] coordinates {
            (0.0, 53.99)
            (0.001, 54.22) (0.005, 54.73) (0.01, 55.37)
		    (0.02, 55.62) (0.04, 55.53) (0.1, 48.1)
	    };

	    \end{axis}
	  \end{tikzpicture}
    }    
    \caption{Can CNN sensitivity be reduced? We test Gaussian filtering and morphological closing of test images before CNN prediction as well as input layer dropout during training.}
    \label{fig:filtered_perturb}
    \end{figure*}

    \myparagraph{Robustness by test image denoising}
    A first idea to handle noisy images is image denoising techniques. 
    We evaluated whether simple linear and non-linear noise removal approaches are suitable to reduce the sensitivity of CNNs.
    \figurename~\ref{fig:filtered_perturb} shows the results for AlexNet in combination with Gaussian filtering and morphological closing on CUB200-2011. 

    Filtering the input images with a Gaussian filter of size $3\times 3$ and $\sigma=0.5$ improved the performance if the variance of the Gaussian RGB noise is greater than 10 or the pepper noise is more frequent than $p=2\%$.
    Using a larger Gaussian filter decreases the accuracy significantly and performs worse even though it leads to increased robustness. 
    The morphological closing operations slightly improves accuracy in case of strong noise, however, the accuracy is worse than a Gaussian filtering with $\sigma=0.5$.
    
    The results reveal that preprocessing improves the robustness, but decreases the accuracy for images without any noise. 
    For noisy test images, the accuracy is higher if noise removal is only subtle. 
    This result is intuitive as noise removal can lead to a loss in information.
    
    \myparagraph{Robustness by augmented training}
    Instead of treating noise at test time, invariance to noise can be also learned at training time.
    Specifically, random noise can be added to the training data during CNN learning. 
    We used a dropout layer~\cite{srivastava2014dropout} added between the input data and the first convolutional layer 
    of the CNN. Dropout indirectly augments the training set by setting neuron outputs to zero at random, which is in our case equivalent of adding noise directly to the images 
    during training.
    In our case, the dropout probability was set to 0.1 and the results are shown in \figurename~\ref{fig:filtered_perturb}.

    Similar to preprocessing the input image, augmented training of a CNN reduces its performance on noise-free test images. 
    However, the performance on noisy images is greatly improved and the CNN is more robust to strong random noise.
    The results suggest that augmented training is beneficial if the test domain is indeed characterized by 
    high degrees of image noise compared to the training domain.
    In any other case, focusing on the training images without any noise augmentation seems to be the better choice.
    
    \myparagraph{Qualitative results}
    The results demonstrate that a large portion of the class predictions change even if only a small amount of noise is added.
    Fig.~\ref{fig:sensitive_examples} shows sample images whose classification score significantly or only marginally changed when noise is added. 
     High sensitivity images are classified by the CNN with high confidence if the image is free of noise, but their maximum classification score changes significantly 
     when even small noise is added (Gaussian RGB noise with $\sigma=1$). Low sensitivity images behave in the opposite way. 
    
    As can be seen in Fig.~\ref{fig:sensitive_examples}, high sensitivity images 
    are characterized by low contrast objects, whereas low sensitivity images
    often have discriminative color patterns.
    To validate this observation, we computed the entropy of hue values 
    within the bird region for the 100 least sensitive and the 100 most sensitive images. Whereas the least sensitive images had a mean hue entropy of $6.12$, 
the most sensitive images were characterized by a mean hue entropy of $5.51$. 
    An additional Wilcoxon rank sum test also showed that this difference in mean
    values is indeed significant ($p<10^{-10}$) and our observation for Fig.~\ref{fig:sensitive_examples} is valid. 

    \newcommand\figh{0.13}
    \begin{figure*}
     \centering
     \textbf{Images with high sensitivity}
    
    \resizebox{\linewidth}{!}{ 
     \includegraphics[height=\figh\linewidth]{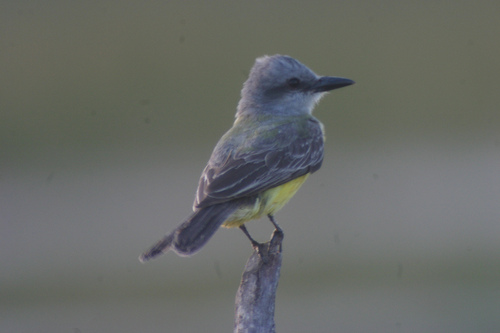}
     \includegraphics[height=\figh\linewidth]{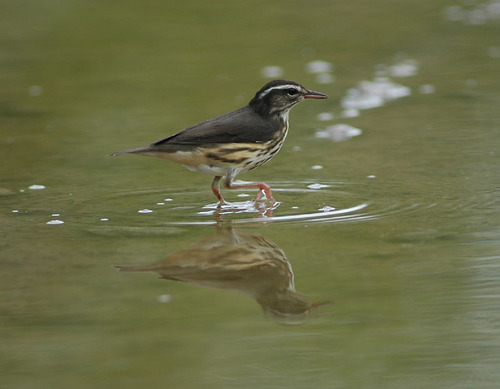}
     \includegraphics[height=\figh\linewidth]{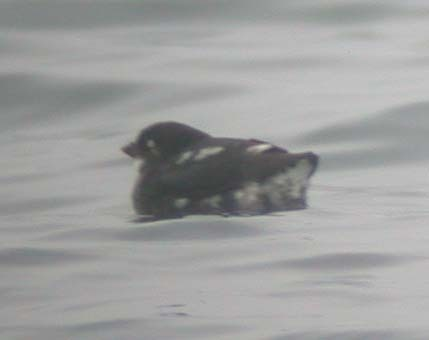}
     \includegraphics[height=\figh\linewidth]{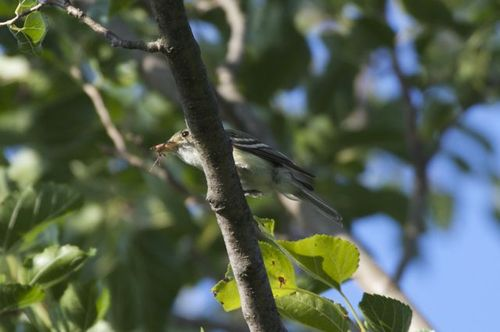}
     \includegraphics[height=\figh\linewidth]{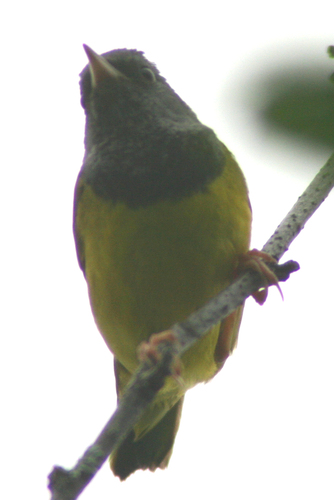}
     \includegraphics[height=\figh\linewidth]{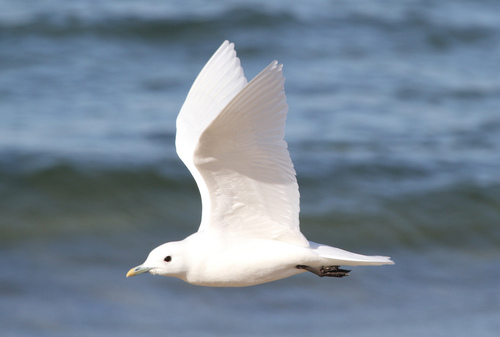}
    }
     
     \textbf{Images with low sensitivity}
 
    \resizebox{\linewidth}{!}{ 
     \includegraphics[height=\figh\linewidth]{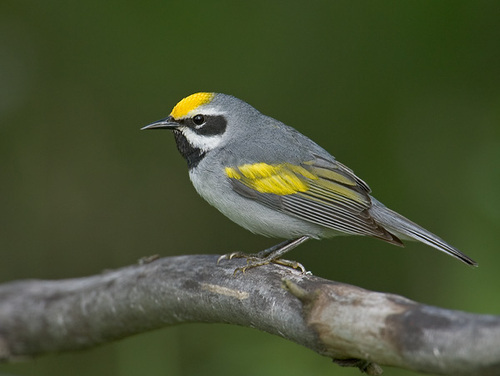}
     \includegraphics[height=\figh\linewidth]{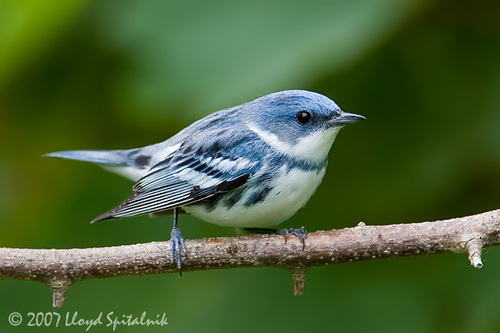}
     \includegraphics[height=\figh\linewidth]{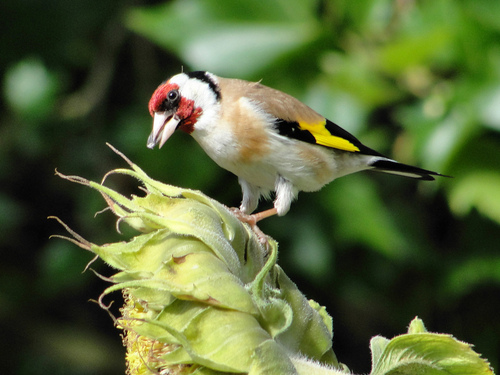}
     \includegraphics[height=\figh\linewidth]{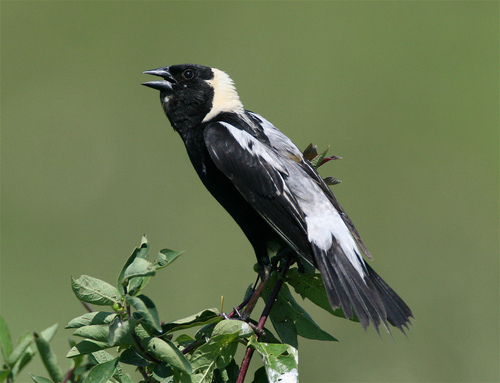}
     \includegraphics[height=\figh\linewidth]{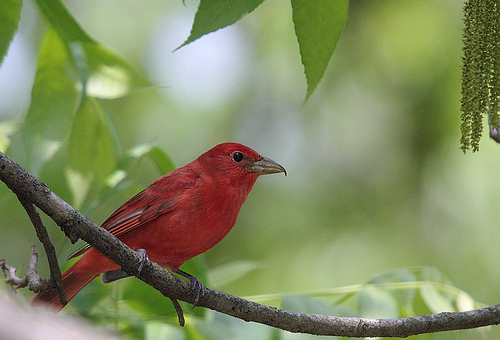}
     \includegraphics[height=\figh\linewidth]{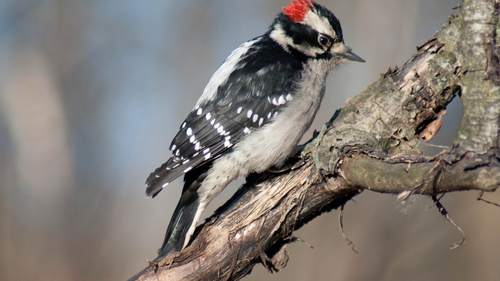}
    }
     \caption{Sample images with a high or low standard deviation of the classification score with respect
        to Gaussian noise ($\sigma=1$). Images have been cropped to ground-truth bounding boxes and scaled
        according to the CNN input layer.}

      \label{fig:sensitive_examples}
     \end{figure*}
    
   \section{Can we predict CNN sensitivity for a test image?}
\label{sec:senspred}

    \newcommand\addfig[1]{\includegraphics[width=0.30\linewidth]{figures/corr_#1.pdf}}
    \begin{figure*}[t]
        \centering
        \addfig{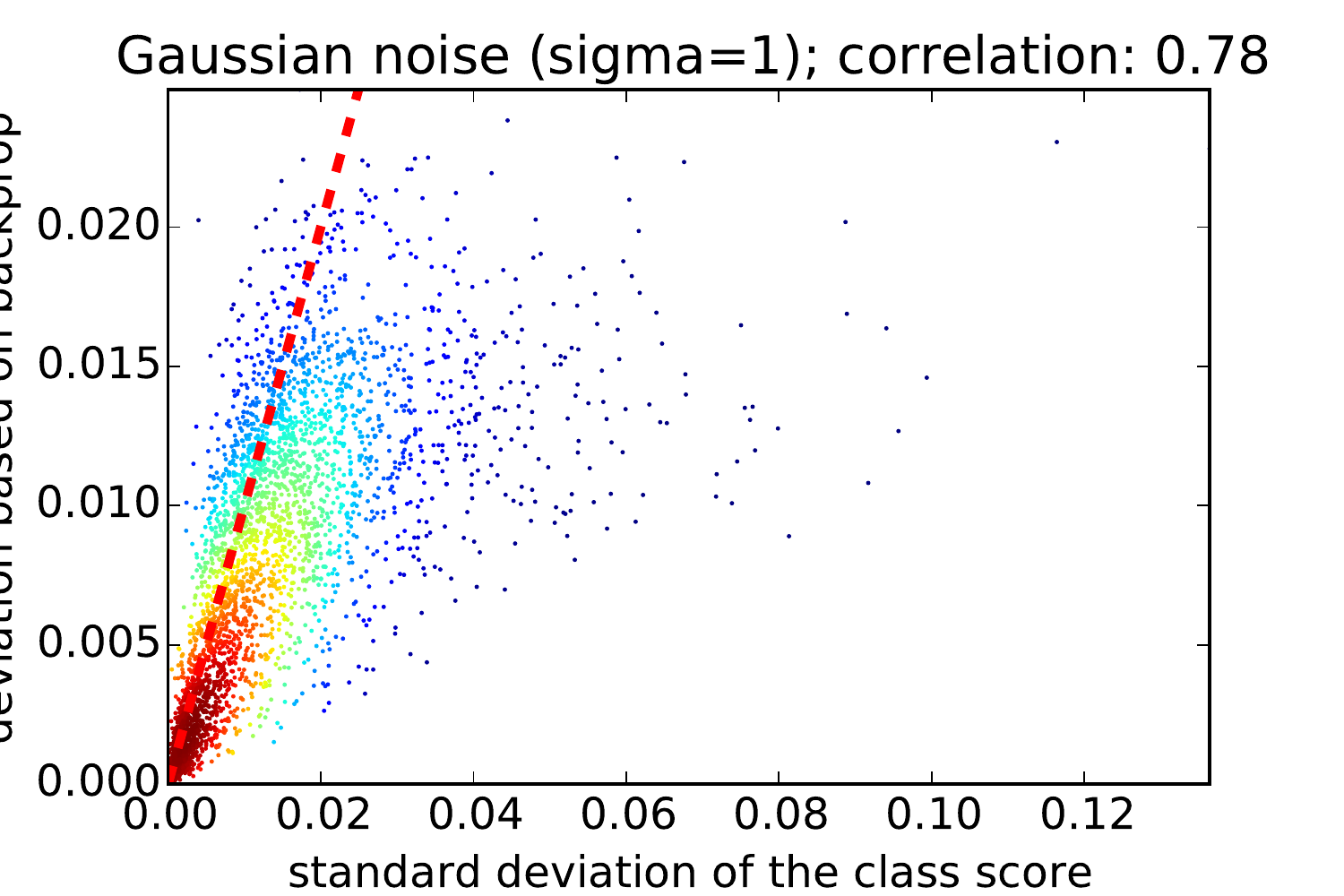}
        \addfig{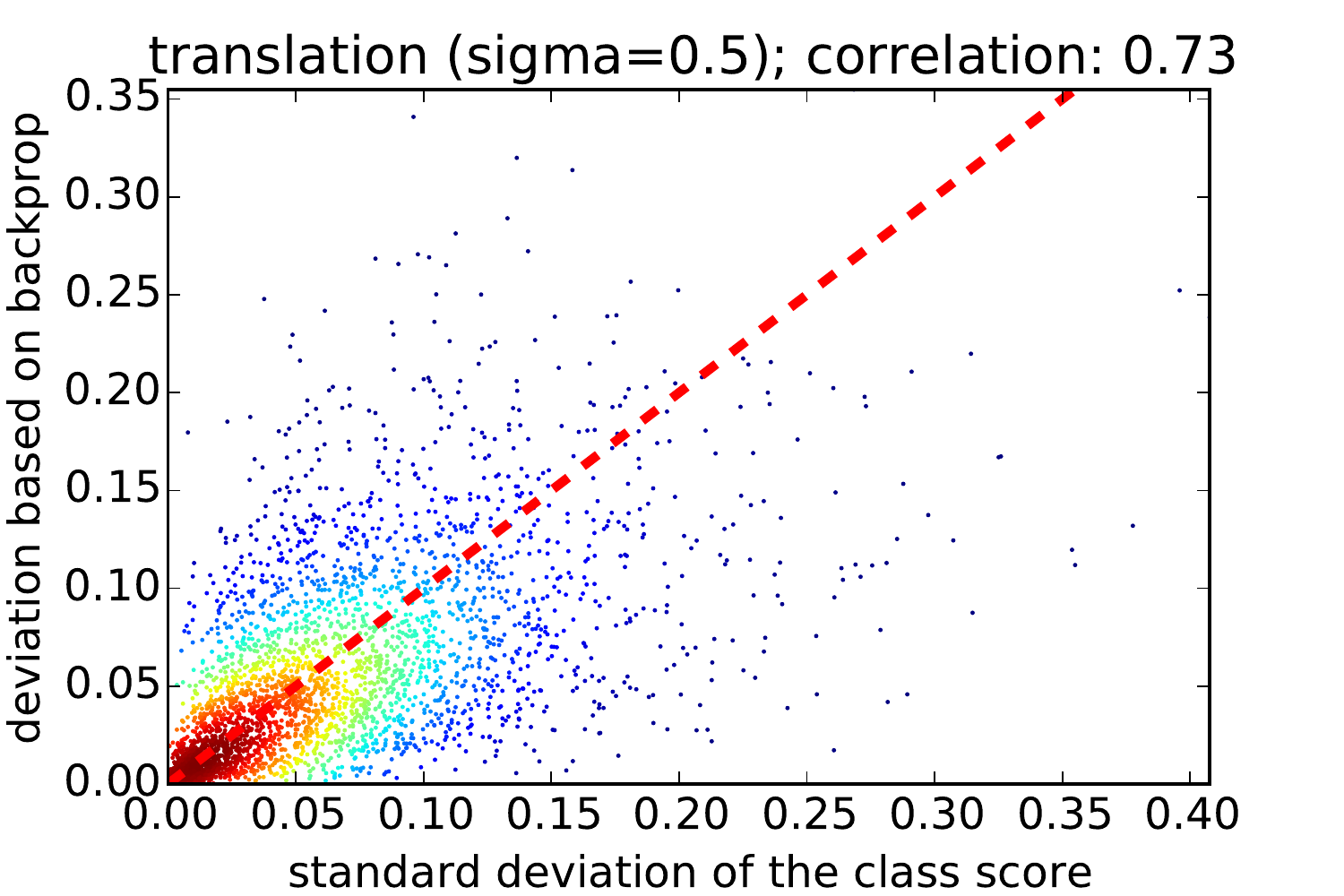}
        \addfig{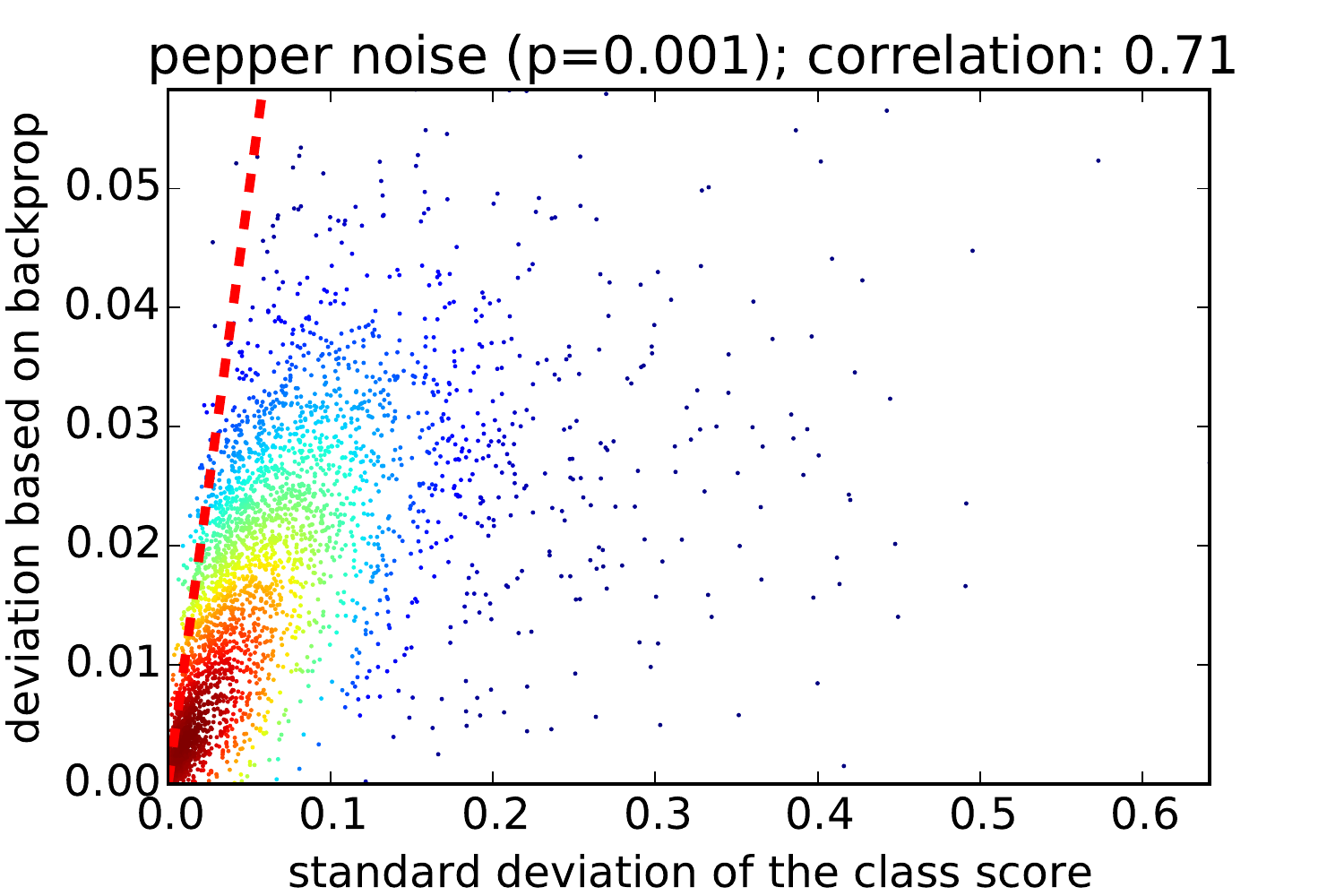}
        \addfig{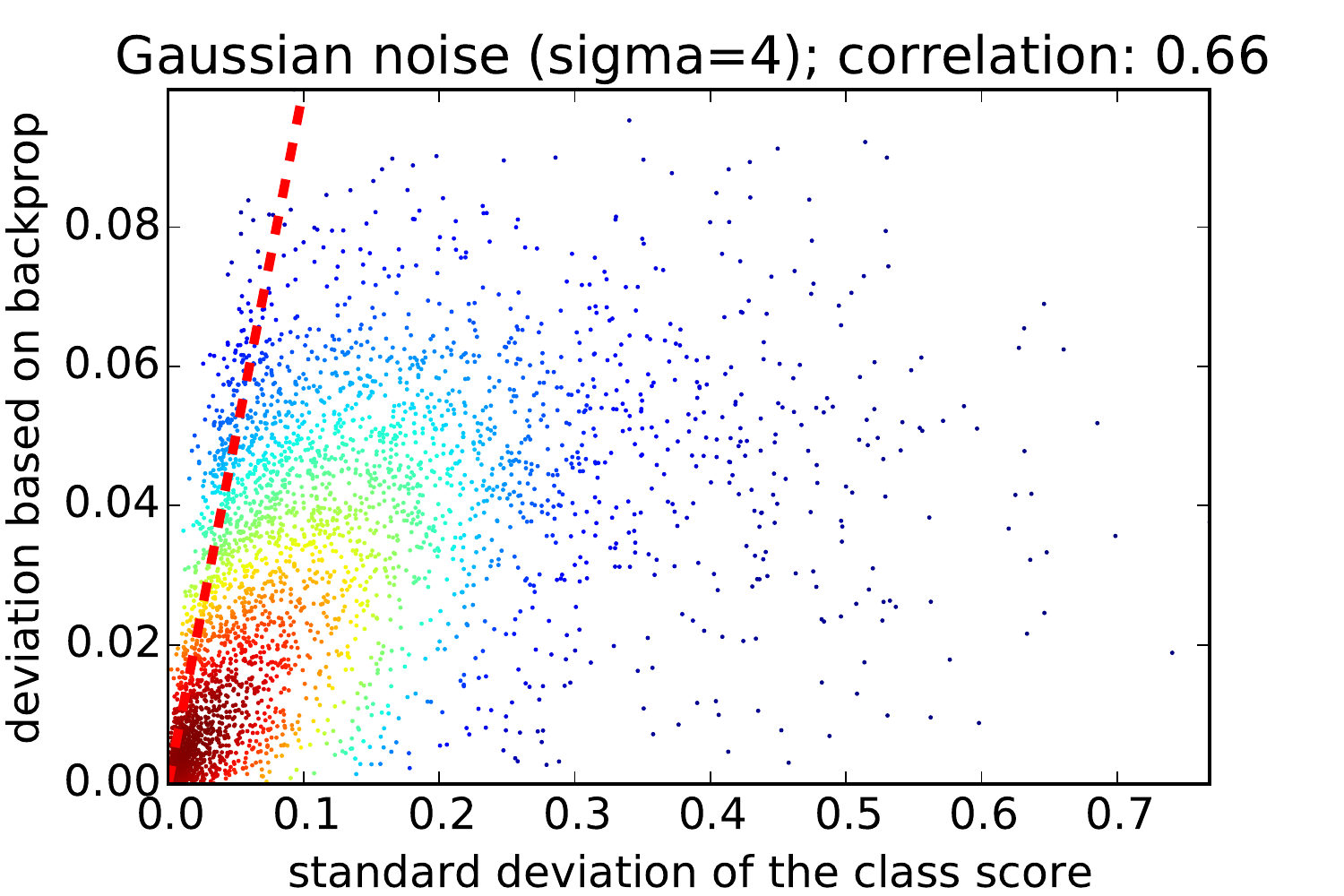}
        \addfig{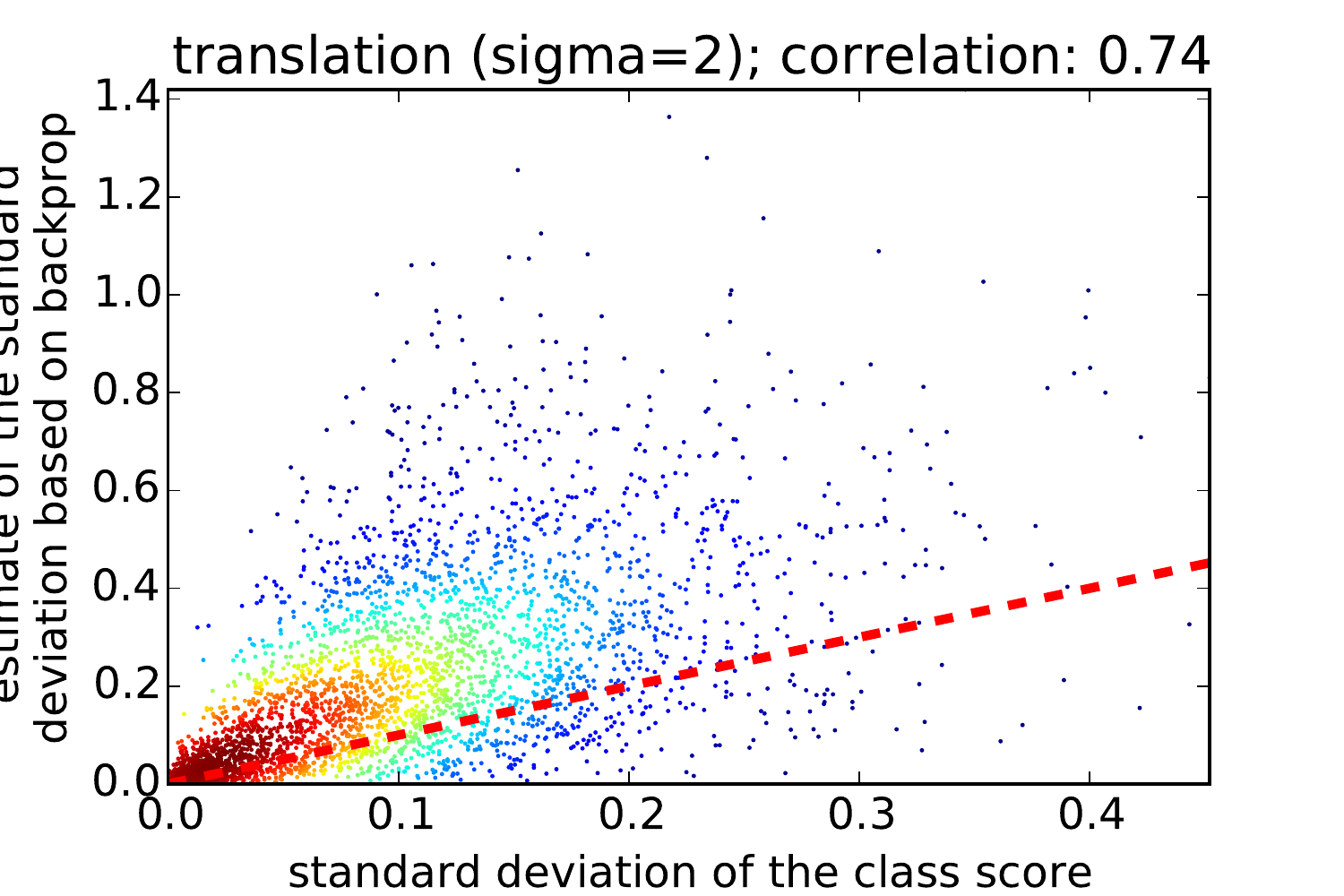}
        \addfig{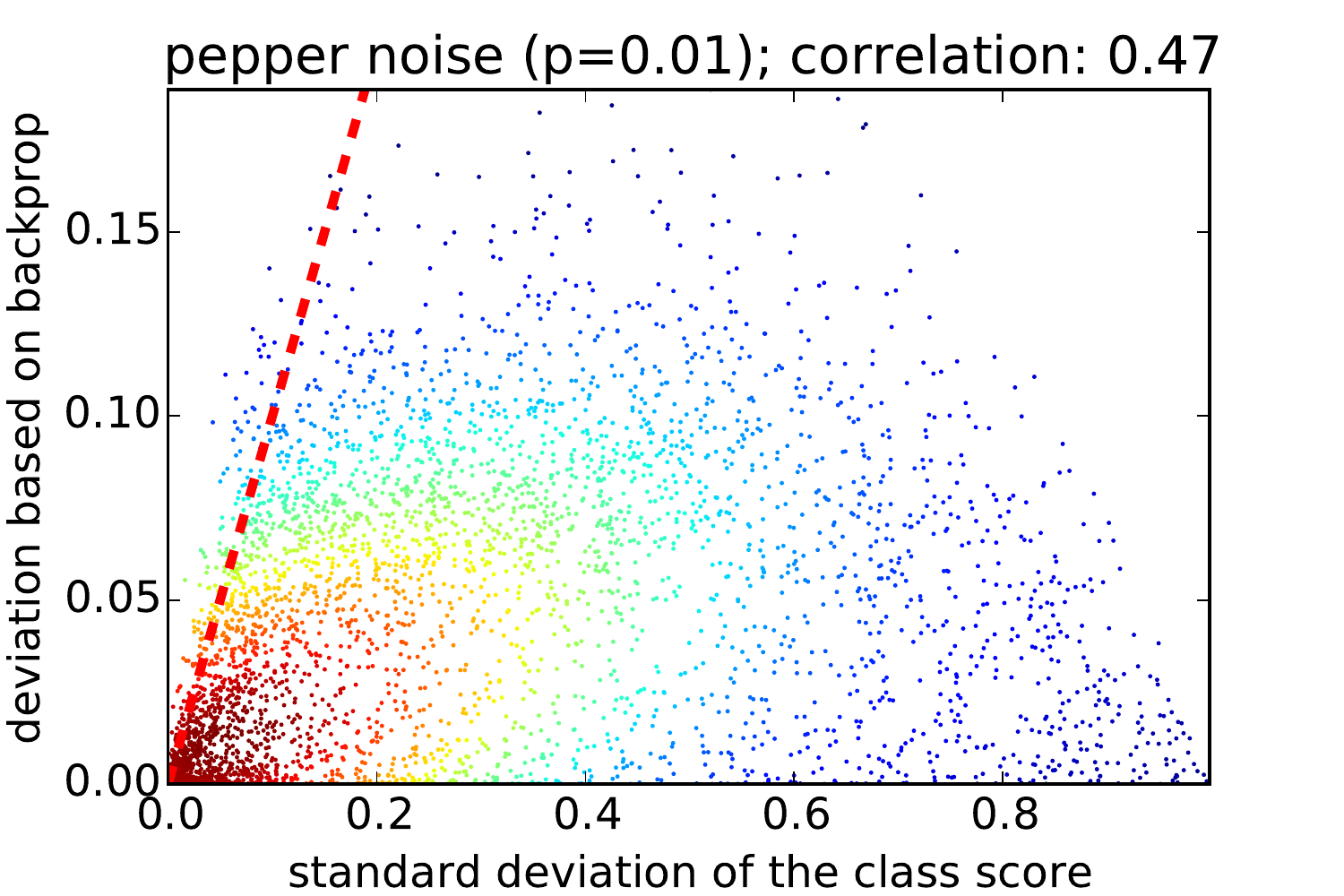}
        \caption{Scatter density plots of our sensitivity prediction for (Left) Gaussian RGB noise with $\sigma \in \{1, 4\}$,
            (Center) translations with $\sigma \in \{0.5, 2\}$, and (Right) pepper noise with $p \in \{0.001, 0.01\}$. 
            Each point in the scatter plot corresponds to a single image in the CUB dataset, which we
            transformed with a given perturbation 10 times. The colors indicate the density in the plot.
            The diagonal red dashed line corresponds to the identity of the estimate from gradients 
            and the empirical estimate from different trials.  
        }
        \label{fig:corr}
    \end{figure*}

    Since we now know that CNN outputs can be sensitive to certain transformations and noise processes, the question
    remains whether we can quickly detect images with unstable CNN outputs. This question goes beyond a pure sensitivity study but
    asks for uncertainty estimates often available for Bayesian methods but not for CNNs.
    In the following, we derive a method for estimating sensitivity scores that does not require a costly explicit alteration of the image.

    \myparagraph{Predicting sensitivity with a backward pass}
    Let $\cnnout(\aimg) \in \mathbb{R}$ be the single output of a CNN for an image $\aimg = (\aimg_1, \ldots, \aimg_C) \in \mathbb{R}^{D \cdot C}$ represented as a flattened vector with $D$ pixels and $C$ channels. 
    In the following, we assume that the input image is altered by $\transform(\aimg, \noiseparam)$,
    where $\noiseparam \in \mathbb{R}^N$ is a random variable controlling the perturbation and
    without loss of generality, we assume that $\transform(\aimg, \bm{0}) = \aimg$.
    We are now considering the change $\triangle \cnnout$ of CNN outputs when applying 
    $\transform$ to an image $\aimg'$ and using a first-order approximation: 
    \begin{align}
        \triangle \cnnout &= \cnnout(\transform(\aimg', \noiseparam')) - \cnnout(\transform(\aimg', \bm{0}))
                          \approx \frac{\partial \cnnout}{\partial \aimg}^T |_{\aimg = \aimg'} \cdot \frac{\partial \transform}{\partial \noiseparam}^T |_{\noiseparam=\bm{0}} \cdot \noiseparam'
                          = \bm{F}^T \cdot \bm{G}^T \cdot \noiseparam'
    \end{align}
    The matrix $\bm{G} \in \mathbb{R}^{N \times C \cdot D}$ is the Jacobian of the perturbation evaluated at $\noiseparam=0$.
    The gradients $\bm{F} = \frac{\partial \cnnout}{\partial \aimg} \in \mathbb{R}^{C \cdot D}$ can be easily computed with a backward pass~\cite{Simon14:PDD,SimonyanSaliency}. 
    They have been used by previous work both for computing saliency and segmentation~\cite{SimonyanSaliency} as well as 
    for part discovery~\cite{Simon14:PDD}. The above result is quite intuitive also from the perspective of saliency maps. If the gradient map shows high saliency values throughout the whole image, the influence of Gaussian RGB noise on the CNN output is likely to be high. This is not the case if only a small part of the image
    is occupied by the object achieving the highest saliency.
    The transformation $\transform$ can be viewed as a stochastic process and we 
    derive an approximation for the variance of the CNN output as follows, where we assume
    that the expectation with respect to $\noiseparam$ is the output for the original image $\aimg'$:
    \begin{align}
        \notag
        \variance_{\noiseparam}(\aimg') &=
        \expectation_{\noiseparam} \left( \cnnout(\transform(\aimg', \noiseparam) - \cnnout(\aimg') ) \right)^2 
        \approx \expectation_{\noiseparam}\left( \noiseparam^T \bm{G} \bm{F} \bm{F}^T \bm{G}^T \noiseparam \right)\\
        &= \text{tr}\left( \expectation_{\noiseparam} (\noiseparam \noiseparam^T) \cdot \bm{G} \bm{F} \bm{F}^T \bm{G}
    ^T \right)
        \label{eq:sensscore}
   \end{align}
    We use $\text{tr}()$ to denote the trace of a given matrix. 
    In the following paragraphs, we study different types of perturbations and
    their resulting gradients $\bm{G}$. We refer to the variance in the above formula
    as the \emph{sensitivity score}.

    \myparagraph{Predicting iid. image noise sensitivity}
    A simple model for perturbations is additive iid.~Gaussian (RGB) noise on the image, which we already analyzed empirically in
Sect.~\ref{sec:expsens}:
$g^{\text{gauss}}(\aimg, \noiseparam) = \aimg + \noiseparam, \quad \noiseparam \sim \mathcal{N}(0, \sigma^2 \bm{I}),$
    where $\noiseparam$ is an $C \cdot D$-dimensional random vector. The Jacobian of the transformation
    is simply the identity matrix, \ie $\bm{G} = \bm{I} \in \mathbb{R}^{C \cdot D \times C \cdot D}$ and
    we obtain:
        $\variance_{\noiseparam}^{\text{gauss}}(\aimg') = \sigma^2 \| \bm{F} \|_2^2$.
    Our sensitivity score is therefore proportional to the magnitude of the gradient map.    

    Let us now consider pepper noise that affects every (RGB) pixel of the image by setting them to black pixels 
    with probability $p$. Following our notation this can be expressed by:
    $
        g^{\text{pepper}}(\aimg, \noiseparam) = \left[ \aimg_1 * (1-\noiseparam), \ldots, \aimg_C * (1-\noiseparam) \right],
    $
    where $\noiseparam \in \{0,1\}^{D}$ is a vector of independent Bernoulli random variables, and $*$ denotes component-wise multiplication.
    The Jacobian $\bm{G}$ is a $\mathbb{R}^{D \times C \cdot D}$ matrix and the multiplication with $\bm{F}$ gives us
    \begin{align}
    \bm{G} \bm{F} = \bigl[ - \sum\limits_{c=1}^{C} \aimgfunc_{k+D \cdot (c-1)} \cdot F_{k+D \cdot (c-1)} \bigr]_{k=1}^{D} = \bm{v} \in \mathbb{R}^{D},
    \end{align}
    a vector of length $D$, which is the number of pixels $D$ of image $\aimg$.
    The matrix $\expectation_{\noiseparam}(\noiseparam \cdot \noiseparam^T)$ depends on the noise probability
    $0 \leq p \leq 1$:
    $\expectation_{\noiseparam}(\noiseparam \cdot \noiseparam^T) = 
        p^2 \bm{e} \bm{e}^T - p (p-1) \bm{I}$,
    with $\bm{e} = \left(1, \ldots, 1\right)^T$. Finally after combining, we have:
    \begin{align}
        \variance_{\noiseparam}^{\text{pepper}}(\aimg') = p^2 ( \bm{v}^T \bm{e} )^2 - p (p-1) \| \bm{v} \|^2.
    \end{align}

    \myparagraph{Predicting image translation sensitivity}

    Following our analysis in Sect.~\ref{sec:expsens}, the gradients with respect to small image translations $\noiseparam \in \mathbb{R}^2$ for all channels are 
    \vspace{-2pt}
    \begin{align}
        \bm{G} &= \begin{bmatrix}
                    \frac{\partial \aimg_1}{\partial x} \ldots \ldots \frac{\partial \aimg_{D\cdot C}}{\partial x} \\
                    \frac{\partial \aimg_1}{\partial y} \ldots \ldots \frac{\partial \aimg_{D\cdot C}}{\partial y}\\
                  \end{bmatrix}
    \end{align}
    \vspace{-2pt}
    and depend on the image gradients in each channel. 
    Since we consider a Gaussian model for the translations $\noiseparam \in \mathcal{N}(\bm{0}, \sigma^2 \bm{I})$,
    we end up with:
        $\variance_{\noiseparam}^{\text{trans}} = \sigma^2 \| \bm{G} \bm{F} \|_{2}^2$.
    where image gradients are multiplied with the CNN gradient map. This result is quite intuitive since large image gradients 
    corresponding to edges will likely lead to a high sensitivity of the CNN with respect to small translations.

    \myparagraph{Validating our sensitivity prediction}
    How accurate is our sensitivity prediction? To answer this question, we correlate our approximations for $\variance$ with the actual empirical standard
    deviation of the output change and provide scatter plots in \figurename~\ref{fig:corr}. A single point in the scatter plot corresponds
    to a single perturbation of one of the CUB test images colored with its density value in the plot to improve visualization of the distribution of data points. CNN outputs are computed using AlexNet.

    As can be seen, the sensitivity prediction is quite accurate for small perturbations with a high correlation (given in the title of each
    figure), which is reasonable since our method is based on a linear approximation. 
    For pepper and Gaussian RGB noise, however, large perturbations lead to a smaller correlation 
    of our sensitivity score with our empirical estimates. This is due to our linear approximation with gradient estimates, which is only valid for smaller noise levels, and the small number of perturbed samples ($10$ in our case) we use for our empirical estimates.

\section{Discussion}
\label{sec:discussion}
The experiments show that the influence especially of common intensity noise is severe even at low noise levels. 
The reason is a domain shift between noise-free training and pertubated test data.
From our study, we can draw several conclusions:
\vspace{-5pt}
\begin{enumerate}
\itemsep-5pt
 \item The training images should have the same noise level as the test images and care has to be taken even for small noise applied to intensities.
 \item Data augmentation during training is not the solution as it decreases the accuracy on noise free images dramatically and is only beneficial for high noise levels as shown.
 \item Noise sensitivity depends on the CNN architecture and VGG19 has shown to be the most robust one.
 \item Sensitivity of CNN outputs can be predicted for small noise levels with our technique in \sectionname~\ref{sec:senspred} allowing for uncertainty estimates of CNN outputs.
\end{enumerate}

\vspace{-5pt}
These conclusions can be seen as guidelines especially for developers of real-world applications, where, for example, cheap camera sensors deliver low quality images but the training was performed on relatively noise-free datasets like ImageNet.
We studied sensor-related perturbations on purpose, since they are often neglected in the era of
huge datasets and benchmarks. Simulating perturbations might not be relevant in scenarios where the sensor used for acquisition
during training is equivalent to the one during testing. In this case, a larger dataset already contains examples of
realistic noise, such as different viewpoints and object deformations. However, our evaluations are important when it
comes to the change to or the use of low-cost sensors with higher noise levels and wrong color calibration.

As it is impossible to simulate all possible real-world perturbations, we only consider a subset of them.
Missing are especially geometric distortions including radial distortion as well as white-balance,
which are hard to realistically simulate. 
The results for translation and global color shift, respectively, lead to the assumption that their influence is fairly low.
Second, multiple kinds of noise occur in combination.
As the addition of noise is likely to decrease accuracy, the combination of multiple noise types might lead to an even stronger degradation. 
For example, Gaussian intensity noise with $\sigma=4$ combined with pepper noise with probability $p=0.02$ yields 48.5\% accuracy on CUB200-2011.
Compared to only Gaussian intensity noise, the accuracy is almost 12\% lower and more than 3\% lower compared to only applying pepper noise.

\section{Conclusions}
\label{sec:conclusions}
In this paper, the sensitivity of common CNN architectures is analyzed empirically and analytically. 
The experiments reveal that especially the most common AlexNet architecture is very sensitive to strong non-adversarial random noise leading to a significant drop in classification performance. 
VGG19 is more robust in terms of stability with respect to Gaussian and pepper noise as well as other non-random perturbations.
Nevertheless, the effect of noise with medium strength already has a significant impact on accuracy in fine-grained recognition tasks.
Two common approaches for dealing with noise were evaluated and turned out to improve stability only for highly degraded images in the test domain.
We further provide an estimation of the sensitivity of a CNN for a given image and show that its predictions strongly correlate with our expensive empirical estimates.

In future work, we plan to study the relationship between architectural choices
in a CNN and its sensitivity to perturbations in more detail, \eg with ResNet-like architectures~\cite{he2015deep}
with multiple depths.

\myparagraph{Acknowledgements} Part of this research
was supported by grant RO 5093/1-1 of the German Research Foundation (DFG).

\begin{todosection}
\section{ToDo, Questions, Discussion Topics}
\begin{enumerate}
        \item Our compensation techniques have been only tested with AlexNet so far.
        \item Sensitivity prediction was only tested with AlexNet.
        \item The sensitivity prediction is missing a real application (CNN combination would be one).
        \item Experiments with the other degradations suggested by Bob (\todo{experiments are running})
        \item Minor: consistent references
        \item Language polishing
\end{enumerate}
\end{todosection}

\bibliography{paper}

\end{document}